\documentclass{amia}
\usepackage{tabularx}
\usepackage{colortbl}
\usepackage{adjustbox} % Include this in the preamble

\usepackage[utf8]{inputenc}
\usepackage{url}
\usepackage{graphicx}
\usepackage{amsmath, amssymb}
\usepackage{enumitem}
\usepackage{multirow}
\usepackage{setspace}
\usepackage{natbib}
\usepackage{etoolbox}
\usepackage{enumitem}
\usepackage{wrapfig}
\usepackage[font=small]{caption}
\usepackage{hyperref}
\usepackage[english]{babel}
\usepackage[table]{xcolor}
\usepackage{bbding}
\usepackage[normalem]{ulem}
\usepackage{comment}
\usepackage{pdfpages}
\usepackage{soul}
\usepackage{xcolor}
\usepackage{nicematrix}
\usepackage{color-edits}
\usepackage{wrapfig}
\usepackage{ragged2e} % For text alignment

 \addauthor[SX]{sx}{teal}

\begin{document}

\title{Toward Large Language Models as a Therapeutic Tool: Comparing Prompting Techniques to Improve GPT-Delivered Problem-Solving Therapy}

%\title{Can GPT help your mental health? \\ A Case Study on GPT's performance in PST Therapy with prompt engineering}

\author{Daniil Filienko, BS, $^1$, Yinzhou Wang, MS, $^3$, Caroline El Jazmi, BS, $^1$, Serena Xie, MS, $^2$, Trevor Cohen, MBChB, Ph.D., $^2$, Martine De Cock, Ph.D.$^1$, Weichao Yuwen, Ph.D., RN$^1$ }

\institutes{
   $^1$University of Washington, Tacoma, WA;
   $^2$University of Washington, Seattle, WA;
   $^3$Dartmouth College, Hanover, NH
}

\maketitle

\section*{Abstract}
\vspace{-1em}
\textit{While Large Language Models (LLMs) are
being quickly adapted to many domains, including healthcare, their strengths and pitfalls remain under-explored. 
In our study, we examine the effects of prompt engineering to guide Large Language Models (LLMs) in delivering parts of a Problem-Solving Therapy (PST) session via text, particularly during the symptom identification and assessment phase for personalized goal setting. We present evaluation results of the models' performances by automatic metrics and experienced medical professionals. We demonstrate that the models' capability to deliver protocolized therapy can be improved with the proper use of prompt engineering methods, albeit with limitations. To our knowledge, this study is among the first to assess the effects of various prompting techniques in enhancing a generalist model's ability to deliver psychotherapy, focusing on overall quality, consistency, and empathy. Exploring LLMs' potential in delivering psychotherapy holds promise with the current shortage of mental health professionals amid significant needs, enhancing the potential utility of AI-based and AI-enhanced care services.
}

\section*{Introduction}
\vspace{-1em}
Numerous studies have demonstrated the potential of Large Language Model (LLM) usage in medical applications, ranging from question-answering services to producing medical notes \cite{nori2023generalist, vanveen2024adapted}. Deployment of LLMs in  AI-supported care services holds the potential to mitigate healthcare costs and broaden access to care.
%Increasing access to such technology could seemingly mitigate healthcare costs
% , alleviating the rising price of healthcare caused by a shortage of medical professionals \cite{shortage} 
%as well as minimize existing inequities in access to healthcare. 
%There already exist companies %like Hippocratic AI 
Companies are already offering products that attempt to undertake the roles of administrative and medical clinicians through the use of LLMs.\footnote{see e.g. https://www.hippocraticai.com/} One of the fields that could greatly benefit from
%in desperate need of 
additional resources is psychotherapy, with as many as 20\% of people worldwide needing mental health care \cite{ee2561611e0045f6be614987f5de6dd2} in the context of a global scarcity of mental health professionals \cite{patel2018lancet}. 
% Traditional online chatbots involve some form of sentiment analysis followed by a templated response based on the user's specific demonstrated needs.

% Here, we adapt a novel technology of Large Language Models to improve upon existing automatic methods for performing psychotherapy.
Previously, studies have demonstrated the advantages of using relevant and empathetic responses in mental health dialogues \cite{mentalsentiment1, Sharma2020-sw}. Althoff et al.~\cite{mentalsentiment1}, in particular, showed that more successful human counselors use fewer templated replies and produce varied responses to similar questions. LLMs' ability to generate coherent and contextually appropriate responses may provide an ideal tool for simulating such behavior, though concerns remain about their safety and utility as patient-facing tools. In this paper, we describe a study of LLMs' ability to provide such relevant and empathetic responses in real-time psychotherapy dialogues. 

In this study, we explore the ability of off-the-shelf LLMs to deliver Problem-Solving Therapy (PST) for family caregivers,  targeting common caregiving symptoms, such as fatigue and anxiety, through a dialogue system that provides caregivers the tools to self-monitor symptoms, problem-solve, and take appropriate actions. 
PST is an effective form of cognitive behavioral therapy with a specific protocol, making it a good test case for LLM performance, and providing the model with specific guidelines to follow.

We demonstrate that non-medical LLMs can display a reasonable performance in PST due to their remarkable ability to use the information provided in the prompt to enhance their performance \cite{fewshot}. We adapt an LLM flow that already showed good performance for medical Q\&A \cite{nori2023generalist} to medical dialogue generation. We explore the extent to which a general-purpose LLM such as GPT-4 can be improved by methods not involving modifying the model's weights and compare it with a previously developed human-curated rule-based system \cite{woz} in the context of PST for family caregivers with standardized patients portrayed by actors. Actors were provided sample responses and narratives describing the personas that they were playing in the interactions with the model to ensure consistent behavior. To compare the LLM performance on this domain-specific task meaningfully, we recruited multiple clinicians to evaluate the dialogues without knowing how the dialogues were created. Hence, our work has two primary objectives. The first is to investigate the use of prompt engineering methods to improve a general-purpose LLM's ability to deliver steps of PST. The second is to provide comprehensive automated and human evaluations of the empathy and overall quality of the models' output.

\textit{\textbf{Related Work}}\\
Existing work to improve the accuracy of LLMs for medical applications leverages (1) \textit{fine-tuning} the model's weights through further training on application-specific data and/or (2) developing novel ways to query the model to trigger a high-quality response by altering the prompt or including relevant information or examples in the prompt, a method called \textit{prompt engineering}. Since for fine-tuning, the end results are a reflection of the quality of the fine-tuning corpora used, effective fine-tuning typically necessitates expensive manual review and in-domain expert curation to produce meaningful improvements in the model's responses \cite{lima}, in addition to highly expensive hardware necessary to run fine-tuning algorithms. In the present study, we focus on prompt engineering, a promising option that has been shown to produce comparable improvements without the resource-intensive fine-tuning \cite{nori2023generalist}.
% We found few studies \cite{wang2021llmempowered_2021, chen2023llmempowered_2023, fu2023llmempoweredenhancing_volunteer} that evaluated the possibility of using LLM in delivering various types of therapy, but differ in the methods used to improve model's performance on the downstream task, using only zero-shot prompting \cite{chen2023llmempowered_2023, fu2023llmempoweredenhancing_volunteer} or performing fine-tuning alone \cite{wang2021llmempowered_2021}.
To our knowledge, this is the first study on improving the accuracy of LLM delivering PST through prompt engineering \cite{wang2021llmempowered_2021}

% \textit{Usage of Large Language Models in Medicine}\\
% With the rise of popularity of AI services, such as ChatGPT from OpenAI, LLM-powered solutions are becoming more prevalent than ever in the industry. LLMs' effectiveness in answering medical questions \cite{nori2023generalist} raised the possibility of more widespread LLM adaption in the medical field. Jo et al., in particular, explored the usage of LLMs in psychological applications and notes the sub-optimal performance of the LLMs, in particular the inability of existing tools to control the model to full mitigate certain biased or harmful behaviors  \cite{llm_psy_survey}, due to the generative nature of LLMs \cite{xu2024hallucination}. In this study, we acknowledge these pitfalls and develop some prompting strategies intended to improve the LLMs' performance.

\textbf{Usage of Large Language Models as Therapeutic Chatbots.}
Literature on the usage of LLMs in psychotherapy is nascent. A few published studies have explored LLMs' ability to lead a therapeutic conversation with a user \cite{chen2023llmempowered_2023,fu2023llmempoweredenhancing_volunteer,wang2021llmempowered_2021}. Fu et al.~described utilizing an LLM to augment a human therapist, providing helpful suggestions rather than singularly leading the conversation, i.e., keeping a human in the loop who can detect hallucinations generated by the LLM and increase the controllability of the system \cite{fu2023llmempoweredenhancing_volunteer}. While such a study provides a setup well-suited for current LLMs, our work explores the current limits of the models' performances in fully autonomous settings to explore the potential for full deployment. Wang et al.~performed fine-tuning of a GPT-2 model to generate suitable PST responses \cite{wang2021llmempowered_2021}. However, their work did not explore prompt engineering methods that recently arose to prominence due to their effectiveness with larger models \cite{nori2023generalist}, producing results comparable to those of fine-tuned models \cite{ovadia2024finetuning}. A recent study~\cite{chen2023llmempowered_2023} explored zero-shot prompt engineering to facilitate a diagnostic conversation based on the Diagnostic and Statistical Manual of Mental Disorders-5 standard \cite{alma9918660570001341}. They explored similar questions to ours, achieving an increase in the model's empathy in responses via the use of zero-shot prompting. 
In this study, we expand on prompt engineering methods by using recently emerged techniques including Zero-Shot \cite{wei2021zeroshotlearners}, Few-Shot \cite{fewshot}, and Zero-Shot Chain-of-Thought \cite{wei2023chainofthought}, which we will refer to as Chain-of-Thought (COT) hereafter. 

% We adapt \cite{nori2023generalist} approach, initially designed for medical question answering, to medical dialogue generation and show that it does indeed improve some of the qualities of the dialogue \ref{sec:Results}.

\textbf{Zero-shot prompting:} in this setup, a model receives only a natural language instruction to perform a task, without prior demonstrations. Relying on the model's pre-trained knowledge, this approach maximizes convenience and potential for robustness, albeit typically being the most ineffective due to the absence of examples that could clarify the task's format or expected output \cite{wei2021zeroshotlearners}. 
% \textit{Role-Playing Zero-Shot Prompting}\\
% Role-playing in zero-shot prompting, where the model is assigned a specific role (e.g., a therapist), lies in its ability to delineate and embed that role within the dialogue, guiding the LLM to adapt its responses according to the designated persona. This approach has been shown to either outperform or match the effectiveness of other prompting techniques, such as Zero-Shot-CoT and Few-Shot-CoT
% \cite{2023kongroleplayprompting}, thereby highlighting its potential reasoning capabilities.
% \textit{Extended Zero-Shot Prompting}\\
% Extended zero-shot prompting, as employed in our methodology, entails incorporating detailed instructions tailored to known specific requirements directly into the prompt. This strategy advances beyond the generality and reusability of standard zero-shot prompts by minimizing the need for responses from the chatbot that could be rendered unnecessary with additional context \cite{white2023promptpatterncatalog}. When specific requirements are known in advance, injecting them within the prompt proves more efficient than relying on the LLM to either solicit further information from the user or resort to making an educated guess regarding the appropriate response \cite{white2023promptpatterncatalog}. This approach refines the interaction process, ensuring a more targeted and effective use of zero-shot prompting by directing the LLM's focus and reducing the likelihood of nondeterministic outcomes.
\textbf{Few-shot prompting:} in few-shot, a model is given a small number of input-output pairs as a guide. This approach enables LLMs to continue generating appropriate output for similar inputs without fine-tuning the model's weights \cite{fewshot}. Optimal example selection is key in few-shot prompting. Brown et al. \cite{fewshot} demonstrated that by incorporating up to eight high-quality examples in prompts, GPT-3's performance on various natural language benchmarks significantly improved \cite{fewshot}. 
% The study underscores the value of precisely choosing a limited set of examples, around eight, to effectively harness few-shot learning for nuanced understanding in contexts like mental health support. 
\textbf{Chain-of-Thought (CoT):} CoT leads LLMs through a step-by-step reasoning process, effectively encouraging them to ``think out loud''. Although the specific style can vary without drastically affecting performance, its presence is crucial for improving problem-solving capabilities \cite{wei2023chainofthought}. This technique encourages the model to consider intermediate tokens, leading to more robust responses. CoT has been shown to significantly boost the problem-solving performance of LLMs across complex tasks, from arithmetic to commonsense challenges \cite{wei2023chainofthought}. 

% In addition, while Fu et al. provided a general schema of their prompt that can be adapted for adjacent use cases \cite{fu2023llmempoweredenhancing_volunteer}, no other study provides a comprehensive list of conceptual techniques that could be used in other types of therapy. Here, we attempt to achieve exactly that, adapting standard prompting techniques that have already demonstrated high effectiveness in the other domains \cite{fewshot} to psychotherapeutic dialogue generation. While our evaluations are PST-specific, our findings have broader implications pertinent to implementing chatbots to deliver other types of psychotherapy. \\
% % Upon looking at current approaches intended to produce LLM model capable of successfully performing psychological therapy, we noticed a gap between prompting techniques discussed in NLP and medical literature. We believe that the former could be improved by incorporating some standard prompting techniques from NLP in the custom prompts crafted by the clinicians, most effectively taking advantage of the both fields' specialties. While our NLP knowledge guided our understanding of 'how' to ask the model for a suitable response, our medical expertise guided us in 'what' should we ask the model to do.

\section*{Methods}
\vspace{-1em}
We used various prompt engineering techniques and their combinations to improve LLM performance when delivering a PST session as a therapist bot. We were guided by findings from Nori et al. who showed that by utilizing various standard prompt engineering techniques, a generalist model, in particular state-of-the-art GPT-4, can respond to medical questions with accuracy on par with models specifically fine-tuned for this task  \cite{nori2023generalist}. We utilize the same methods and examine whether their insights can be applied to psychotherapy. While in this study we used GPT-4 through Microsoft Azure's Application Programming Interface (API), at the time of writing, GPT-4 is also available to a wide audience as the backend of OpenAI's ChatGPT conversational agent.\footnote{https://chat.openai.com/} 

In preliminary experiments, our team evaluated multiple prompting strategies in delivering PST and found large inconsistencies in expected model behavior. Models produced at times low quality or generally incoherent responses when performing full PST. To mitigate this, we decided to focus on the two PST steps that were the hardest to control and were most prone to producing unexpected and potentially harmful dialogue outputs. The two steps are \textit{problem selection and identification} (in our case scenario for family caregivers, we named it \textit{caregiving symptom assessment}) and \textit{goal setting}. By merging our technical insights in prompt engineering with clinicians' knowledge of PST, we designed prompts that are specifically relevant to these two steps of PST. As Van Veen et al.~\cite{vanveen2024adapted} have previously shown, human assessments of LLM performance do not always closely correlate with automated metrics, especially in domains requiring expert knowledge, such as healthcare. Hence, in addition to using an automated empathy evaluation method developed by Sharma et al.~\cite{Sharma2020-sw} to evaluate the models' responses, we had a group of clinicians familiar with PST evaluate our dialogues without knowledge of whether the dialogue was LLM-derived. We provide more details about the prompt design and the evaluation process below.

\textbf{\textit{Designing the prompts}} \label{approaches} \\
We started with a naive prompt shown in Table \ref{tab:prompts} and used it as a baseline. We then included the three prompt engineering techniques. We also experimented with combinations of these techniques for better downstream performance.  
\vspace{-1em}
\begin{table*} [ht]
    \centering
    \begin{tabular}{p{0.15\linewidth} | p{0.82\linewidth}}
     \hline
     \textbf{Type of Prompt} & \textbf{Content of the Prompt}\\ 
     \hline
     Baseline & 
     Your responsibility is to guide the conversation with a caregiver through the principles of PST to improve one significant symptom the caregiver is experiencing. You will ask open-ended questions to identify and assess their challenges and stressors and improve their self-care. Avoid focusing on the care receiver. Remember, your job is to help the caregiver.  
     
  When the caregiver asks for goal suggestions by saying `Can you suggest some goals for me?', take this as your cue to thoroughly review the conversation you've had with them. Concentrate on identifying their unique needs and aspirations as discussed. After this review, generate two concise, achievable, and personalized goals that directly address and support their expressed needs and aspirations. Ensure these goals are not only realistic but designed to inspire and boost the caregiver's motivation.\\
     \hline
     Zero-Shot &
     In the process of identifying and assessing the caregiver's symptoms, assess all five aspects if it has not been mentioned in the conversation: \newline 1. Symptom Identification (What are the caregiver's symptoms?) \newline 2. Symptom Frequency (How often do the caregiver's symptoms occur?) \newline 3. Symptom Context (Where and when do the caregiver's symptoms occur? Are there specific people present, or are certain activities involved?) \newline 4. Symptom Severity (On a scale of 1 to 5, how severe are the caregiver's symptoms?) \newline 5. Previous Measures (What has the caregiver already tried to alleviate the symptoms?)\\
     \hline
     Few-Shot &
     Below are ideal dialogue examples illustrating how you, the assistant, should evaluate and address the challenges and stressors of caregivers, referred to here as the user, during conversations. \newline **Attached would be 9 excerpts from multi-turn PST therapy with a human therapist**\\
     \hline
     Chain-of-Thought &
     Think about the user's input step by step.\\
     \hline
    \end{tabular}
    \caption{Our prompts categorized by various types of prompting techniques used}
    \label{tab:prompts}
\end{table*}

\textbf{Baseline:} 
Leveraging insights from the RoleLLM framework \cite{wang2024rolellm}, which highlights substantial improvements in language model performance through role-conditioned instruction tuning, we started by implementing a role-playing approach to augment our chatbot's functionality as a PST assistant. This foundational prompt within our system prompt architecture, shown in Table \ref{tab:prompts}, is composed of two parts which serve two distinct functions. First, it defines the chatbot’s role as an assistant to caregivers, delineating its core functions, tasks, and desired behaviors to facilitate conversations aligned with PST principles, aiming at the identification and analysis of caregivers' symptoms. Second, it ensures proper behavior of the model during the goal setting stage, encouraging it to provide proper goals.
\textbf{Zero-Shot:} To enhance the chatbot's ability to generate structured dialogue, following the PST protocol, we built upon the baseline prompt with the instructions shown in the second row of Table \ref{tab:prompts}, reflecting the sub-stages necessary for accomplishing symptom identification and assessment, as well as the goal suggestion steps of PST.
%
% This initiative leverages the Flipped Interaction Pattern \cite{white2023promptpatterncatalog}, aiming to transition the chatbot’s interactions from user-initiated Question and Answering dialogue, where the LLM tends to simply answer user's questions, to a more proactive PST style therapy, where chatbot would be proactively guiding the conversation to meet specified PST assessment goals.
%
% In order to further improve not just overall dialogue flow, but PST-specific symptom identification step, we provided a more detailed zero-shot propmt. 
Central to our strategy is including precise language explaining five predefined criteria necessary for the successful identification and assessment of caregiver symptoms: Symptom Identification, Symptom Frequency, Symptom Context, Symptom Severity, and Previous Measures \cite{woz}. 
%
% Specifically, if any of the five critical aspects have not been adequately addressed during the interaction, the chatbot is instructed to engage with targeted questions related to these aspects. 
%
This approach is an attempt to enhance the depth and relevance of the chatbot's interactions, grounding its dialogue framework in these five essential aspects of the symptom identification and analysis step in PST. 
\textbf{Few-Shot:} Our implementation of the few-shot prompting method incorporates nine high-quality examples curated by clinicians as in-context learning examples to capture the complexities of symptom identification and analysis within the PST framework \cite{yuwen2022comparingcaregivingneeds}. These examples aim to demonstrate to the model a set of ideal responses to realistic user inputs, providing 
% Despite exceeding the number of examples recommended by previous research \cite{fewshot},
%Our approach ensures 
comprehensive coverage of the symptom identification and analysis phase. %referred to as Step 3 at "A Wizard-of-Oz Interface and Persona-Based Methodology for Collecting Health Counseling Dialog" \cite{kearns2020woz}. 
The first six examples present single-turn dialogues addressing the 5 sub-steps %3.1 to 3.5 
of PST that are listed in the zero-shot prompt (see the second row in Table \ref{tab:prompts}), spanning %a broad spectrum of 
symptom identification and assessment, including frequency, context, severity, and prior mitigation efforts. Additional examples featuring multi-turn dialogues are designed to enhance the chatbot’s capability to conduct coherent and contextually relevant conversations within PST. The multi-turn dialogue examples are essential for demonstrating to the chatbot the intricacies of ongoing therapeutic dialogue, including the need for follow-up questions and the reiteration of the points made to the user during the dialogue.
\textbf{Chain-of-Thought:} Our implementation of CoT is an attempt to systematically improve the chatbot's analytical and problem-solving capabilities by forcing it to be more `thoughtful' in its style of output, 
which can lead to better outcomes as demonstrated in previous studies \cite{nori2023generalist,wei2021zeroshotlearners}. 

% \begin{table}[ht]
% \centering
% \begin{tabular}{|l|l|>{\centering\arraybackslash}m{3cm}|>{\centering\arraybackslash}m{3cm}|}
% \hline
% \textbf{Model} & \textbf{Prompt Structure}                                      & \textbf{Preliminary} \textbf{Assessment} & \textbf{Evaluation} \\ \hline
% 0                 & Baseline                                 & Pass & Proceed \\ \hline
% \rowcolor{gray!50} 1                 
% & Baseline + Zero-Shot                            & Fail & N/A \\ \hline
% 2                 & Baseline + Few-Shot                           & Pass & Proceed \\ \hline
% 3                 & Baseline + Zero-Shot + Few-Shot               & Pass & Proceed \\ \hline
% 4                 & Baseline + Zero-Shot + Few-Shot + CoT    & Pass & Proceed \\ \hline
% 5                 & Baseline + Few-Shot + CoT              & Pass & Proceed \\ \hline
% \rowcolor{gray!50} 6
% & Baseline + Zero-Shot + CoT                & Fail & N/A \\ \hline
% 7                 & Human-curated rule-based system& Pass & Proceed \\ \hline
% \end{tabular}
% \caption{Overview of Model Structures}
% \label{tab:model_structures}
% \end{table}

\begin{table}[ht]
\centering
\begin{tabularx}{\textwidth}{|l|X|>{\centering\arraybackslash}m{2.5cm}|>{\centering\arraybackslash}m{2.5cm}|}
\hline
\textbf{Model} & \textbf{Prompt Structure} & \textbf{Preliminary Assessment} & \textbf{Evaluation} \\ \hline
0 & Baseline & Pass & Proceed \\ \hline
\rowcolor{gray!50} 1 & Baseline + Zero-Shot & Fail & N/A \\ \hline
2 & Baseline + Few-Shot & Pass & Proceed \\ \hline
3 & Baseline + Zero-Shot + Few-Shot & Pass & Proceed \\ \hline
4 & Baseline + Zero-Shot + Few-Shot + CoT & Pass & Proceed \\ \hline
5 & Baseline + Few-Shot + CoT & Pass & Proceed \\ \hline
\rowcolor{gray!50} 6 & Baseline + Zero-Shot + CoT & Fail & N/A \\ \hline
7 & Human-curated rule-based system & Pass & Proceed \\ \hline
\end{tabularx}
\caption{Overview of Model Structures}
\label{tab:model_structures}
\end{table}

\textbf{\textit{Model Development}} \label{approaches} \\
We developed the models gradually, each step aimed at mitigating the faults noticed at the previous stage (Table 2). We produced models of various complexity, which we then evaluated for their ability to achieve the intended objectives.

\textbf{Model 0:} A baseline model, referred to as Model 0, introduces the chatbot in its role of guiding caregivers through step 3, "Identify and Assess Symptom," and step 4, "Goal Setting," following the principles of PST.
\textbf{Model 1 and Model 2:} Expanding on Model 0, Models 1 and 2 improve the chatbot's ability to follow PST structure by introducing key prompt engineering techniques. Model 1 builds upon Model 0 by integrating a set of structured guidelines into the prompt via zero-shot prompting. This approach systematically guides the chatbot into assessing caregiver's challenges and stressors in a specific order. Model 2 further advances Model 0 through the integration of few-shot learning, which leverages selected single-turn and multi-turn example dialogues from previously recorded sessions to set guidelines for the conversation flow. 
\textbf{Model 3 through 6:} In our further explorations, inspired by the methodology of Nori et al.~\cite{nori2023generalist}, we utilize various combinations of prompt engineering techniques across Models 3 to 6. Model 3 integrates zero-shot's structured symptom assessment with few-shot learning to produce more robust output that more closely follows the intended response style. Model 4 incorporates CoT prompting to improve the models' problem-solving results. Models 5 and 6 adjust these components to explore their combined effects on the chatbot's PST performance, with a focus on more effective goal setting. \textbf{Model 7:} While not a model per se, we also curated four dialogues from the rule-based chatbot baseline to compare against our methods following our previous work \cite{woz}. 

\begin{table*}[h]
\centering
\fontsize{10}{12}\selectfont
\begin{tabular}{|l|l|l|}
\hline
Dimensions & Statements or Questions & Measurements \\
\hline
Symptom assessment & \begin{tabular}[c]{@{}l@{}}The therapist successfully assessed the \\ five aspects of the caregiver's symptoms.\end{tabular} & \multirow{2}{*}{\begin{tabular}[c]{@{}l@{}}5-point Likert scale \\ (1 = Completely disagree;\\ 5 = Completely agree)\end{tabular}} \\
\cline{1-2}
Goal setting & \begin{tabular}[c]{@{}l@{}}The goal suggested by the therapist is \\ appropriate for the caregiver.\end{tabular} & \\
\hline
Emotional Reactions & \begin{tabular}[c]{@{}l@{}}The therapist expressed emotions such as \\ warmth, compassion, concern, or similar \\ feelings towards the caregivers.\end{tabular} & \multirow{3}{*}{\begin{tabular}[c]{@{}l@{}}3-point Likert scale \\ (0 - therapist not expressing them at all; \\ 1 - therapist expressing them to \\ some weak degree; \\ 2 - therapist expressing them strongly.)\\ We used the framework from \\ Sharma et al. \cite{Sharma2020-sw} to evaluate peer-to-peer \\ dialogues for both algorithmic \\ and human evaluation.\end{tabular}} \\
\cline{1-2}
Interpretations & \begin{tabular}[c]{@{}l@{}}The therapist communicated an \\ understanding of feelings and \\ experiences inferred from the caregiver's \\ responses.\end{tabular} & \\
\cline{1-2}
Explorations & \begin{tabular}[c]{@{}l@{}}The therapist improved their \\ understanding of the caregiver \\ by exploring the feelings and \\ experiences not stated in the \\ caregiver's response.\end{tabular} & \\
\hline
Overall & Overall, how was this therapy session? & \begin{tabular}[c]{@{}l@{}}5-point Likert scale \\ (1 - very bad; \\ 5 - very good)\end{tabular} \\
\hline
\end{tabular}
\caption{Evaluation components}
\label{tab:eval_metrics_tab}
\end{table*}

\textbf{\textit{Generating Dialogues for Evaluation}} \label{approaches} \\
To achieve a higher level of consistency and protect caregivers' privacy, we used personas to generate dialogues. We first crafted four caregiver personas with exemplary replies based on our prior work with family caregivers. We recruited three research team members with direct professional experiences interacting with family caregivers. They were asked to portray one or two of the personas and converse with each bot therapist (Model 0 through 6 in Table \ref{tab:model_structures}). They were instructed to use consistent and exemplary replies as much as possible. We collected 28 PST dialogues (4 personas * 7 models). We randomized the sequence of the models and did not disclose which specific models they interacted with. After we collected all the dialogues, the research team examined the conversations and found that Models 1 and 6 had at least one dialogue in which the bot therapist asked all five symptom assessment questions at once. When the caregiver responded to one of the questions, the bot therapist could not recognize that one question had been answered and repeated all the questions. This behavior was undesired and would automatically receive low ratings in the human evaluation. Thus, we eliminated them in this preliminary step prior to human evaluation. The models that proceeded to human evaluation were Models 0, 2, 3, 4, and 5. In addition, the research team curated a set of four dialogues with human-delivered PST sessions. These dialogues were curated based on the four personas and the research team's prior rule-based bot-delivered PST protocol \cite{woz}. In total, we had 24 dialogues (6 models * 4 personas) that proceeded to human evaluation. 

\textbf{\textit{Evaluating Dialogues - Human Evaluation}} \label{approaches} \\
We recruited seven clinicians, namely four nurses and three clinical psychologists, to evaluate the quality of the therapy responses. The evaluators were unaware of the models and evaluated the dialogues in random sequence. Evaluators evaluated each dialogue on two aspects, conversational quality and perception of relational capacity, which were adopted from the chatbot evaluation mechanisms by Zhang et al.~\cite{Zhang2020-se}. For conversational quality, since the generated dialogues specifically focused on the ``symptom assessment'' and ``goal setting'' steps of PST, the evaluators were asked to evaluate the quality of these two steps. To evaluate the symptom assessment step, we asked the evaluators to consider if the therapist assessed all five aspects of symptoms. To evaluate the goal-setting step, evaluators were asked to assess the appropriateness of the goals suggested by the therapist in the dialogue. For the perception of relational capacity, we focused on therapist empathy, which was shown to be a predictor for therapy outcome \cite{Elliott2018-yr}.  We employed the three communication mechanisms to measure empathy developed by Sharma et al. \cite{Sharma2020-sw}: Emotional Reactions (ER), Interpretations (IN), and Explorations (EX). Strong empathetic communication expresses emotions reacting to what the user said (Emotional Reaction), communicates an understanding of the user's feeling or experience (Interpretation), and explores the user's feelings and experiences that are not stated (Exploration) \cite{Sharma2020-sw}. Evaluators evaluated all three aspects.  Additionally, we asked evaluators to provide a rating based on the overall impression of the therapy session. Details about the evaluation components and questions are included in Table \ref{tab:eval_metrics_tab}. Moreover, to gain a deeper understanding of the ratings and what factors contributed to the ratings, we also asked the evaluators to provide a brief rationale for each rating.

\textbf{\textit{Evaluating Dialogues - Automatic Evaluation}} \label{automatic_eval} \\
In addition to human evaluation, we adapted the algorithm from Sharma et al. \cite{Sharma2020-sw} to rate the conversations' empathy. We used the default implementation of the algorithm provided on the authors' GitHub repository, a base RoBERTa classifier trained on Reddit dialogue data labeled by the authors. The classifier quantifies the model's 
Emotional Reactions (ER), measuring its ability to express positive emotions when responding to a user's post;
Interpretations (IP), evaluating the model's ability to produce a relevant shared experience;
and Explorations (EX), which captures a level of active interest and engagement with the user's post. 
% The results seemed to be generally consistent with our manual evaluations.

\textbf{\textit{Data Analysis}} \label{summary_eval} \\
Each of the six models that generated dialogues received 28 sets of ratings (4 persona-based dialogues * 7 evaluators). We computed the averages and standard deviations of the ratings for each model. We report on the models with the highest and lowest ratings on each evaluation component. Three team members conducted a rapid deductive analysis of the qualitative rationales provided by the evaluators to identify specific factors contributing to the ratings. 

\setlength{\tabcolsep}{0.85pt}
\section*{Results} \label{sec:Results}
\vspace{-1em}

% Quantitative Ratings
\renewcommand{\arraystretch}{1.3} %  vertical padding
\setlength{\tabcolsep}{3pt} %  horizontal padding
\begin{table}[H]
\centering
\resizebox{\columnwidth}{!}{%
\begin{tabular}{|l|l|l|ll|ll|ll|l|}
\hline
\multicolumn{1}{|c|}{}                       & \multicolumn{1}{c|}{\begin{tabular}[c]{@{}c@{}}Symptom \\ Assessment\\
(1-5)\\\end{tabular}} & \multicolumn{1}{c|}{\begin{tabular}[c]{@{}c@{}}Goal \\ Setting\\
(1-5)\\\end{tabular}} & \multicolumn{2}{c|}{\begin{tabular}[c]{@{}c@{}}Emotional \\ Reactions\\
(0-2)\\\end{tabular}} & \multicolumn{2}{c|}{Interpretations (0-2)}                              & \multicolumn{2}{c|}{Explorations (0-2)}                                 & \multicolumn{1}{c|}{Overall (1-5)} \\ \cline{2-10} 
\multicolumn{1}{|c|}{\multirow{1}{*}{Model}} & \multicolumn{1}{c|}{Human}                                                         & \multicolumn{1}{c|}{Human}                                                   & \multicolumn{1}{c|}{Human}                & \multicolumn{1}{c|}{Algorithm}          & \multicolumn{1}{c|}{Human}       & \multicolumn{1}{c|}{Algorithm} & \multicolumn{1}{c|}{Human}       & \multicolumn{1}{c|}{Algorithm} & \multicolumn{1}{c|}{Human}   \\ \cline{2-10} 
\multicolumn{1}{|l|}{0}                       & 3.15 (1.35)                                                                        & 4.00 (1.09)                                                                  & \multicolumn{1}{l|}{1.56 (0.58)}          & 0.83 (0.17)                             & \multicolumn{1}{l|}{1.59 (0.64)} & 0.04 (0.06)                    & \multicolumn{1}{l|}{1.50 (0.69)} & 0.79 (0.21)                    & 3.68 (1.06)                  \\ \hline
2                                            & 4.68 (0.55)                                                                        & 4.74 (0.45)                                                                  & \multicolumn{1}{l|}{1.71 (0.46)}          & 0.93 (0.08)                             & \multicolumn{1}{l|}{1.64 (0.49)} & 0.06 (0.10)                    & \multicolumn{1}{l|}{1.39 (0.74)} & 1.22 (0.08)                    & 4.36 (0.73)                  \\ \hline
3                                            & 4.75 (0.44) & 4.46 (0.74)                                                                  & \multicolumn{1}{l|}{1.57 (0.57)}          & 0.92 (0.19)                             & \multicolumn{1}{l|}{1.68 (0.55)} & 0.11 (0.02)                    & \multicolumn{1}{l|}{1.21 (0.74)} & 1.29 (0.27)                    & 3.82 (0.90)                  \\ \hline
4                                            & 4.80 (0.41)                                                                        & 4.25 (0.93)                                                                  & \multicolumn{1}{l|}{1.68 (0.48)}          & 0.91 (0.18)                             & \multicolumn{1}{l|}{1.75 (0.44)} & 0.02 (0.03)                    & \multicolumn{1}{l|}{1.32 (0.61)} & 1.31 (0.16)                    & 3.82 (0.79)                  \\ \hline
5                                            & 3.70 (1.35)                                                                        & 3.86 (1.38)                                                                  & \multicolumn{1}{l|}{1.64 (0.49)}          & 0.93 (0.25)                             & \multicolumn{1}{l|}{1.71 (0.46)} & 0 (0)                          & \multicolumn{1}{l|}{1.35 (0.69)} & 0.76 (0.18)                    & 3.61 (1.13)                  \\ \hline
7                                            & 4.62 (0.5)                                                                       & 3.59 (1.01)                                                                  & \multicolumn{1}{l|}{1.22 (0.80)}          & 0.89 (0.14)                             & \multicolumn{1}{l|}{1.22 (0.89)} & 0.04 (0.06)                    & \multicolumn{1}{l|}{0.77 (0.65)} & 1.31 (0.13)                    & 3.19 (1.18)                  \\ \hline
\end{tabular}%
}
\caption{Evaluation results. The results are reported in Mean (Standard Deviation) format.}
\label{tab:eval_results_tab}
\end{table}

Table 4 shows the mean scores with standard deviations of both the human and automated evaluations. In general, Models 2, 3, and 4 were rated higher compared to the human-curated rule-based system (Model 7) in all evaluated areas and delivered the two steps of PST successfully. Overall, evaluators rated Model 2 the best, which is also the highest-rated model for goal-setting and the exploration dimension of empathy. The human-curated rule-based model was rated the lowest across all evaluated aspects except for symptom assessment.
\begin{table*}[h]
\centering
\renewcommand{\arraystretch}{0.99} % Decrease space between lines
\begin{tabular}{|l|l|l|l|}
\hline
\multicolumn{1}{|c|}{\begin{tabular}[c]{@{}c@{}}Evaluation \\ Aspect\end{tabular}} & \multicolumn{1}{c|}{\begin{tabular}[c]{@{}c@{}}Human \\ Rating\end{tabular}} & \multicolumn{1}{c|}{Model} & \multicolumn{1}{c|}{\begin{tabular}[c]{@{}c@{}}Example Chatbot Response \\\end{tabular}} \\ \hline
\multirow{2}{*}{\begin{tabular}[c]{@{}l@{}}Symptom \\ Assessment\end{tabular}} & \begin{tabular}[c]{@{}l@{}}Highest \\ (mean=4.80)\end{tabular} & 4 & \begin{tabular}[c]{@{}l@{}}“It sounds like you're dealing with some intense emotions. I want to acknow-\\ ledge how challenging it can be to manage fear and sadness, especially in a \\ caregiving role. To better understand and help you with these feelings, could \\ you tell me how often you experience these feelings of fear and sadness?”\end{tabular} \\ \cline{2-4} 
 & \begin{tabular}[c]{@{}l@{}}Lowest \\ (mean=3.15)\end{tabular} & 0 & \begin{tabular}[c]{@{}l@{}}“It sounds like you're carrying a heavy load both emotionally and practically. \\ Let's address these one at a time. Firstly, regarding emotional support, can \\ you tell me about your current support system? Do you have friends, family, \\ or a community you can talk to about what you're going through? Secondly, \\ for the practical help with caregiving tasks, have you been able to identify \\ specific tasks that are particularly challenging or time-consuming? Are there \\ routine aspects of care that you think someone else could assist you with?”\end{tabular} \\ \hline
\multirow{2}{*}{\begin{tabular}[c]{@{}l@{}}Goal \\ Setting\end{tabular}} & \begin{tabular}[c]{@{}l@{}}Highest \\ (mean=4.74)\end{tabular} & 2 & \begin{tabular}[c]{@{}l@{}}“Excellent. Since you're open to online support groups and you're also looking \\ to enhance your emotional support, here are two personalized  goals that might \\ be beneficial for you: 1. **Join an Online Support Group**: Within the next \\ week, aim to identify and join at least one online support group for caregivers. \\ Spend a few minutes each day, or as your schedule allows, to engage with the \\ group. This can be a place where you can share your experiences, receive \\ advice, and connect with others who understand what you're going through.\\ 2. **Schedule Regular Check-ins**: Set up a weekly check-in with a friend or \\ family member who can offer emotional support. This could be a phone call, \\ video chat, or even a text exchange, where you can share your feelings and \\ challenges. Having a set time each week for emotional check-ins can provide \\ you with something to look forward to and ensure you have a consistent outlet \\ for your emotions. How do these goals sound to you? Are they something you \\ feel you can commit to?”\end{tabular} \\ \cline{2-4} 
 & \begin{tabular}[c]{@{}l@{}}Lowest \\ (mean=3.59)\end{tabular} & 7 & \begin{tabular}[c]{@{}l@{}}“Thanks for being open to it. Which of the two would you like to try for this \\ week? 1. Journal writing. \\ 2. Practice the 4-7-8 deep breathing exercise technique .”\end{tabular} \\ \hline
\end{tabular}%
\caption{Symptom Assessment and Goal Setting Examples}
\label{tab:symptom_goal}
\end{table*}

\textbf{Symptom assessment.} 
For symptom assessment, the improved models (Models 2-5) were all able to assess a caregiver's symptom in at least four turns of dialogue. When the ``caregiver'' shared a challenge such as \textit{``I'm feeling overwhelmed caring for my father with cancer and looking for support,''} the bot therapist would first provide an empathetic response, followed by a therapeutic component that continued the PST process. For example, one response by a bot to this specific ``caregiver'' was, \textit{``Hello, I'm really glad you reached out for support. It's incredibly important to take care of yourself, especially when you're in a caregiving role. It sounds like you're going through a lot right now. Can you tell me more about what specifically is making you feel overwhelmed?''}  The average rating for symptom assessment across the four models was 4.5 out of 5, indicating that ``the therapist successfully assessed the five aspects of the caregiver's symptoms". Model 4, with all prompt engineering techniques, scored the highest for symptom assessment (mean = 4.80). The majority of the evaluators thought Model 4 successfully assessed all aspects of the caregiver's symptoms and had good symptom identification. Model 0 using only baseline prompting was rated the lowest. Evaluators reported that this bot missed assessing certain aspects of caregivers' symptoms, such as severity and frequency, and had poor follow-up questions on context. Moreover, several evaluators mentioned that the bot asked multiple questions at a time and made it overwhelming. Tables 5 and 6 show example bot responses from the models with the highest and lowest rating for each criterion, including symptom assessment. 

\textbf{Goal setting.}
When suggesting health goals for the caregivers to improve an identified symptom, Model 2 was rated the best (mean = 4.74). Evaluators' feedback on Model 2 included keywords ``personalized",  ``collaborative", ``realistic", ``appropriate", ``specific", ``straightforward", ``reasonable", ``relevant", ``well-structured", and ``detailed". Model 7 (i.e. human-curated rule-based model) was rated the lowest (mean=3.59). Evaluators pointed out a few drawbacks including being ``not clear", ``not specific", ``superficial", and ``providing only one suggestion". Sample chatbot responses from Model 2 and Model 7 for goal setting are included in Table 5. In the same persona mentioned above, one bot therapist made the following goal recommendation: \textit{``Let's work on setting some goals that can help you manage these overwhelming feelings more effectively. Based on what you've shared, here are two personalized goals that you might consider: 1. Establish a Daily Unwinding Ritual: Given that your feelings of fear and sadness intensify during the evenings and nights, creating a calming evening routine could be beneficial. This might include activities like meditation, listening to soothing music, or practicing gentle yoga. The goal would be to engage in this ritual every evening for at least 15 minutes to help transition into a more relaxed state of mind before bedtime. 2. Seek Emotional Support: It's important to have a support system in place. You could set a goal to reach out to a friend, family member, or a support group for caregivers at least once a week. Sharing your experiences and feelings with others who understand can provide comfort and may offer new perspectives or coping strategies. How do these goals sound to you?''}

\begin{table*}[h]
\centering
\renewcommand{\arraystretch}{0.99} % Decrease space between lines
\begin{tabular}{|p{0.15\linewidth}|p{0.12\linewidth}|p{0.03\linewidth}|p{0.60\linewidth}|}
\hline
\multicolumn{1}{|c|}{\begin{tabular}[c]{@{}c@{}}Evaluation \\ Aspect\end{tabular}} & \multicolumn{1}{c|}{\begin{tabular}[c]{@{}c@{}}Human \\ Rating\end{tabular}} & \multicolumn{1}{c|}{Model} & \multicolumn{1}{c|}{\begin{tabular}[c]{@{}c@{}}Example Chatbot Response \\ (Algorithm rating on a scale of 0-2)\end{tabular}} \\ \hline
\multirow{2}{*}{\begin{tabular}[c]{@{}l@{}}Emotional \\ Reactions\end{tabular}} & \begin{tabular}[c]{@{}l@{}}Highest \\ (mean=1.71)\end{tabular} & 2 & \begin{tabular}[c]{@{}l@{}}“That must be quite exhausting for you. Not getting enough sleep can \\ certainly exacerbate stress and pain. To better understand your \\ situation, can you tell me more about the context of your insomnia? \\ For example, are there specific thoughts or worries that keep you \\ awake, or is it the pain that's more disruptive?” (Algorithm rating: 1)\end{tabular} \\ \cline{2-4} 
 & \begin{tabular}[c]{@{}l@{}}Lowest \\ (mean=1.22)\end{tabular} & 7 & \begin{tabular}[c]{@{}l@{}}“I can see that. I have one more question to ask about your \\ overwhelmed feeling. Have you done anything to help you with your \\ overwhelmed feeling?” (Algorithm rating: 1)\end{tabular} \\ \hline
\multirow{2}{*}{Interpretations} & \begin{tabular}[c]{@{}l@{}}Highest \\ (mean=1.75)\end{tabular} & 4 & \begin{tabular}[c]{@{}l@{}}“That sounds really tough, being at home and feeling that heightened \\ awareness of your daughter's anxiety and the challenges in connecting \\ with her.” (Algorithm rating: 0)\end{tabular} \\ \cline{2-4} 
 & \begin{tabular}[c]{@{}l@{}}Lowest \\ (mean=1.22)\end{tabular} & 7 & \begin{tabular}[c]{@{}l@{}}“I hear you. Let’s use the Problem Solving Therapy (PST) process that \\ I mentioned to you to explore how we may address this.” \\ (Algorithm rating: 0)\end{tabular} \\ \hline
\multirow{2}{*}{Explorations} & \begin{tabular}[c]{@{}l@{}}Highest \\ (mean=1.50)\end{tabular} & 0 & \begin{tabular}[c]{@{}l@{}}“That sounds quite challenging and must be affecting your daily \\ functioning. Not being able to sleep well for 4-6 nights a week is \\ significant. Can you share more about the context of these sleepless \\ nights? For example, are there specific thoughts related to your \\ daughter’s illness or your pain that tend to keep you awake?” \\ (Algorithm rating: 2)\end{tabular} \\ \cline{2-4} 
 & \begin{tabular}[c]{@{}l@{}}Lowest \\ (mean=0.77)\end{tabular} & 7 & \begin{tabular}[c]{@{}l@{}}“Great. Let’s use the Problem Solving Therapy (PST) process that I \\ mentioned to you to explore how we may address this.” \\ (Algorithm rating: 2)\end{tabular} \\ \hline
\end{tabular}%
\caption{Emotional Reactions, Interpretations, and Explorations Examples}
\label{tab:emotional_interpretations_explorations}
\end{table*}

\textbf{Empathy.}
For the three dimensions of empathy, the human-curated rule-based model was rated consistently as the lowest across all three dimensions, with other models being rated similarly. Model 2 scored the highest in ER (mean = 1.71), followed by Model 4 (mean = 1.68). One evaluator expressed approval of the emotional reactions exhibited by the therapist bot powered by Model 2, stating \textit{``I think the empathy came across better in this one than some others because it made a statement and then asked a question. It didn't pile questions on top of questions like in some of the prompts... it makes the conversation seem less empathetic and two sided."} 
Model 4 was rated the highest in the IP (mean = 1.75). The majority of the evaluators expressed that there was good use of empathetic statements and reflections. One evaluator provided an example of ``an excellent follow-up on feeling/experiences and insights into the feelings" in which the bot responded \textit{``It can be harder at night when it's quiet."} For the Exploration dimension of empathy, evaluators rated Model 0 the highest (mean = 1.50). One evaluator expressed that \textit{``In this session, the therapist does more open ended questions that get to more of the root of the problem."} For the lowest rated human-curated rule-based model across all three dimensions of empathy, evaluators reported that the bot said things like \textit{``got it''} that demonstrated poor empathy and regurgitated sentences like \textit{``I am so sorry to hear that ..."}, which made it feel robotic. Several evaluators also reported that the human-curated rule-based model displayed ``zero or very limited explorations of deeper thoughts and feelings." Moreover, the bot was perceived as ``fake and too cheerful." One evaluator said that \textit{``It used the word `glad' very often and also thanked the caregiver quite often. It's important for the bot to in some ways mirror the caregiver, meet them where they are at. And someone worrying 4 out of 5 may not want it to seem overly cheerful."} According to the automated evaluation on empathy, all models performed very similarly in terms of ER, with an average close to 1, meaning that the therapist expressed emotional reactions to some weak degree. Rarely were our models nor the human dialogue able to demonstrate Interpretations as all models were scored close to 0 by the  automated evaluation algorithm. The majority of the models, except for Model 0 and Model 5, demonstrated moderate empathy in Explorations, with Models 4 and the human-curated rule-based model scored the highest (mean = 1.31) by the automated evaluation.
\section*{Discussion}
\vspace{-1em}
%write a simple summary of what we did and what we found - 3-4 sentences
% The two models with baseline and few-shot (with and without CoT) seemed to perform better in most of the evaluation items. However, it was difficult to identify one model that performed the best across evaluation aspects. This could be due to the complexity of psychotherapy tasks as compared to classification or QA answering, requiring in-domain expertise to be properly assessed, and various professionals may evaluate the examples vastly differently.
% \sxcomment{limitation:}While demonstrating better following of PST-specific guidelines and improved performance overall, the model still suffers from fundamental flow of being uncontrollable and at times producing unexpected results even with low generation temperature \cite{xu2024hallucination}, making it more deterministic. Even after specific guidelines being provided, we cannot guarantee that the model will always follow them without producing some harmful input. In general, we noticed that prompt engineering seems to improve the performance of the model over the naive baseline, however there exists a lot of space for improving its performance for the application of the psychotherapy in general and PST in particular by exploring post-training tools that modify the weights of the model, such as performing fine-tuning \cite{ovadia2024finetuning} or continually aligning the model to human feedback via Reinforcement Learning from Human Feedback or Direct Preference Optimization. 
In this study, we used various prompting techniques to improve GPT-delivered PST for family caregivers and used both human and automatic algorithms to evaluate the therapy dialogues. We found that by using prompt engineering techniques, we were able to improve the quality of the therapy conversations beyond the baseline prompt, but with considerable limitations. Empathy evaluated by both human and the algorithm did not vary significantly across models, despite improving over the baseline in the emotional reaction and interpretation dimensions. 
% This novel study contributed to the literature by examining LLMs' potential in delivering complex psychotherapy.
% Psychotherapy is a nuanced procedural task that has a wide set of implicit and explicit guidelines that are expected to be followed when performed on real patients. It is not dependent on content extraction abilities alone and does not usually have one best answer like Q\&A \cite{nori2023generalist}, but is dependent on the LLM's ability to generate both safe content and use proper style.
% Hence, coupled with human evaluation, our study allows for a more robust understanding of LLM's generational abilities.
Below, we discussed specific prompt engineering techniques and their performances. %However, the difference between the manual and automatic evaluations of LLMs in general is yet to be properly quantified, and we leave that for future work. 

%From our methods, 
Zero-shot learning did not perform as well as the other techniques that we evaluated. Our finding that explicit directions may not be sufficient to adapt a model to a domain-specific task is in line with the literature \cite{nori2023generalist,fewshot}. %In fact, some zero-shot only methods were deemed so inadequate in our preliminary evaluation, that we decided not to proceed with human evaluation of those methods. %to save human resource, because of the method's obvious incapability to guide PST therapy. 
In our scenario, the difference in performance between zero-shot learning and the other techniques was more pronounced because of the nature of the downstream task, i.e.~PST being a protocolized therapy. PST requires a specific way of conversing with the user. Zero-shot prompting focuses on explicitly defining tasks for the model to follow, but many aspects of what constitutes good therapy are implicit (e.g.~more actionable advice being preferred over more generic, overly optimistic comments being preferred less \cite{Sharma2020-sw}). The possible approaches to achieve the explicit goals are almost endless. %Only a very small subset of them fit both external and internal guidelines. Here, the proverb stands true that it is better to see something once (few-shot learning) than to hear about it a thousand times because examples provide both explicit and implicit content and style standards that the model can further imitate. 
Explicitly demonstrating examples of high-quality PST responses as part of few-shot prompt led to better model performance, showing that providing the model with data is still necessary to produce coherent dialogues. 
%Both of the models that were filtered out at the preliminary evaluation were models without few-shot technique. 
%We suggest that it is primarily because few-shot examples permit the model to capture the style of existing therapy instances. We think that 
Our findings align with previous results from the literature, that show that providing few-shot examples typically noticeably increases the models' performance compared to zero-shot baselines on tasks involving the generation of coherent word sequences, %(or in other words, word sequences following a proper in-domain style), such as in task of Translation or Using Novel Words, 
akin to how PST demands a coherent dialogue \cite{fewshot}.

Adding CoT resulted in better empathy, especially exploration. However, it reduced the quality of symptom identification and goal setting. The original CoT work by Wei et al.~\cite{wei2023chainofthought} did not explore dialogue-generating tasks, and it is possible that we could experiment with alternative CoT prompts that would improve empathy while maintaining task quality. However, given that LLMs are fundamentally next-word predictors trained on large corpora of text, it is possible that in the pre-training data of the model, chunks of text beginning with phrases similar to our prompt happened to be more exploratory and less actionable. We find that explanation reasonable because, in human writings, texts nudging someone to think tend to be more contemplative than action-driven in nature. 

%However, to provide a more robust reasoning for that phenomena, we would need to do a full ablation study, but since we focused on improving LLM's ability to conduct therapy, rather than developing a science of prompting, providing reasons behind LLMs' behaviors, it is outside the scope of our work. We think it could be a valuable work, since currently there exist no true explanation of the LLMs' output tendencies, with the users relying on heuristics and experience, rather than some known prompting theories. 
% It took me couple hours to understand that it is not a human baseline. - it's not an ideal baseline. I'm writing this paragraph right now to explain it. It was from a Woz study, but the Woz study's intention was to use human to teach bot to deliver the therapy. 
% Thank you for explanation 

In line with previous work \cite{vanveen2024adapted}, we found low agreement between the human and automatic evaluations. This may be partially due to the model by Sharma et al. \cite{Sharma2020-sw} trained on single-turn Reddit posts, presenting a different setting than PST. Future studies should examine the generalizability of the automatic evaluation algorithms with domain shifts.

Although the LLM-delivered therapy dialogues received higher scores than the human-curated baseline, it is worth noting that the human baseline is not actual therapist-delivered PST dialogues. We adapted dialogues the team created for developing a rule-based chatbot to deliver PST to family caregivers. Some responses were designed to be generic such as using \textit{``I am sorry to hear that''} instead of a more empathetic response using psychotherapeutic techniques such as normalizing (e.g., \textit{``I am sorry to hear that. Many caregivers feel isolated while caring for an ill child.''}). In our study, to protect real caregivers' privacy, we used a persona-based approach to collect dialogues. It would be preferable to engage with actual caregivers. In the future, we plan to develop privacy-preserving technologies in order to guarantee the privacy of the user's information when training or interacting with LLMs, so that our in-context learning processes do not expose users' private information to commercially-hosted LLMs. 
% \textbf{\textit{Future Directions}} \label{future_directions} \\
% Our next works' directions is to 
% \begin{enumerate}
This study intended to explore the ability of off-the-shelf LLMs with prompt engineering to deliver part of PST and achieved promising results. In the future, we will continue to improve the models to deliver full PST by performing fine-tuning and Retrieval Augmented Generation (RAG) techniques. In the aforementioned reading comprehension task \cite{fewshot}, we can see that fine-tuned models still can outperform generalists with few shot examples. However, recent studies \cite{nori2023generalist, ovadia2024finetuning} showed that it is not always true, possibly due to the increased knowledge contained in the larger models, allowing them to perform on-par with or better than fine-tuned models. We will continue to explore the best techniques or combinations of techniques that enable an automated chatbot to deliver a full PST session. 
    % \item Design new explainability mechanisms that can allow us to quantify the certainty of the Large Language Models and potentially mitigate the model's hallucinations. While being a novel research area, we believe it is necessary for applications involving high-risk users, such as those utilizing the therapy bots. Evaluate the ability of fine-tuning to outperform prompt engineering alone. While fine-tuning greatly depends on the fine-tuning corpora quality \cite{lima}, studies shown that is still is a viable, albeit more resource intensive, alternative to prompt engineering \cite{ovadia2024finetuning, fewshot}.
% \end{enumerate}

% Also, concluding with a list of future directions makes the work seem unfinished. Better would be a sentence after each related limitation that explains how this might be addressed in future work.

\section*{Conclusion}
\vspace{-1em}
In this paper, we adapt multiple novel prompt engineering approaches to improve an LLM's ability to deliver part of a psychotherapy session. Consistent with previous findings \cite{chen2023llmempowered_2023}, we demonstrate that the model's capability to deliver protocolized therapy can be improved with the proper use of prompt engineering methods, albeit with limitations. Through both automatic and human evaluation, we show an improvement over the baseline model after applying our methods to PST, demonstrating that some prompt engineering techniques are better at improving the performance of the models than others. Hence, while the current models cannot be deployed directly in psychotherapy settings without human oversight, this work contributes to the effort in exploring the potential of LLMs as a therapeutic tool. As such, this work represents an important step toward using LLMs to address the limited availability of human therapists in the context of an escalating need for mental health services. 

\textbf{Acknowledgments.} The authors would like to thank Microsoft for the UW Azure Cloud Computing Credits for Research program. Daniil Filienko is a Carwein-Andrews Distinguished Fellow.
This research was, in part, funded by the National Institutes of Health (NIH) Agreement No.1OT2OD032581 and R21NR020634, and the Rita and Alex Hilman Foundation Emergant Innovation Grant. The views and conclusions contained in this document are those of the authors and should not be interpreted as representing the official policies, either expressed or implied, of the NIH.
\vspace{-1em}

% References as numbers
\makeatletter
\renewcommand{\@biblabel}[1]{\hfill #1.}
\makeatother
% Eliminate space between entries and set to single space
\setlength{\bibsep}{0.0pt}
\singlespacing

\bibliographystyle{vancouver}
\bibliography{amia}  

\end{document}